\documentstyle[aclap]{article}

\title{\vspace{-0.5in}Evaluation of Semantic Clusters}
\author{Rajeev Agarwal\\
Mississippi State University\\
Mississippi State, MS 39762\\
USA\\
rajeev@cs.msstate.edu\\}

\begin{document}

\maketitle
\vspace{-0.5in}
\begin{abstract}

Semantic clusters of a domain form an important feature that can
be useful for performing syntactic and semantic disambiguation.  Several
attempts have been made to extract the semantic clusters of a domain by
probabilistic or taxonomic techniques.  However, not much progress has been
made in evaluating the obtained semantic clusters.  This paper
focuses on an evaluation mechanism that can be used to evaluate semantic
clusters produced by a system against those provided by human
experts.

\end{abstract}

\section{Introduction\footnote{The author is currently at Texas Instruments
and all inquiries should be addressed to rajeev@csc.ti.com.}}

Most natural language processing (NLP) systems are designed to work on certain
specific domains and porting them to other domains is often a very
time-consuming and human-intensive process.  As the need for applying NLP
systems to more and varied domains grows, it becomes increasingly important
that some techniques be used to make these systems more portable.  Several
researchers
\cite{LanHir-88,RauJacZer-89,Pus-92,GriSte-93,BasPazVel-94b}, either
directly or indirectly, have addressed issues that assist in making it easier
to move an NLP system from one domain to another.  One of the reasons for the
lack of portability is the need for domain-specific semantic features that
such systems often use for lexical, syntactic, and semantic disambiguation.
One such feature is the knowledge of the semantic clusters in a domain.

Since semantic classes are often domain-specific, their automatic acquisition
is not trivial.  Such classes can be derived either by distributional means or
from existing taxonomies, knowledge bases, dictionaries, thesauruses, and so
on.  A prime example of the latter is WordNet which has been used to provide
such semantic classes \cite{Res-93b,BasPazVel-94b} to assist in text
understanding.  Our efforts to obtain such semantic clusters with limited
human intervention have been described elsewhere \cite{Aga-95}.  This paper
concentrates on the aspect of evaluating the obtained clusters against classes
provided by human experts.

\section{The Need}

Although there has been a lot of work done in extracting semantic classes of a
given domain, relatively little attention has been paid to the task of
evaluating the generated classes.  In the absence of an evaluation scheme, the
only way to decide if the semantic classes produced by a system are
``reasonable'' or not is by having an expert analyze them by inspection.  Such
informal evaluations make it very difficult to compare one set of classes
against another and are also not very reliable estimates of the quality of a
set of classes.  It is clear that a formal evaluation scheme would be of great
help.

Hatzivassiloglou and McKeown (1993) cluster adjectives into partitions and
present an interesting evaluation to compare the generated adjective classes
against those provided by an expert.  Their evaluation scheme bases the
comparison between two classes on the presence or absence of pairs of words in
them.  Their approach involves filling in a YES--NO contingency table based on
whether a pair of words (adjectives, in their case) is classified in the same
class by the human expert and by the system.  This method works very well for
partitions.  However, if it is used to evaluate sets of classes where the
classes may be potentially overlapping, their technique yields a weaker
measure since the same word pair could possibly be present in more than one
class.

An ideal scheme used to evaluate semantic classes should be able to handle
overlapping classes (as opposed to partitions) as well as hierarchies.
The technique proposed by Hatzivassiloglou and McKeown does not do a good job
of evaluating either of these.  In this paper, we present an evaluation
methodology which makes it possible to properly evaluate overlapping classes.
Our scheme is also capable of incorporating hierarchies provided by an expert
into the evaluation, but still lacks the ability to compare hierarchies
against hierarchies.

In the discussion that follows, the word ``clustering'' is used to refer to
the set of classes that may be either provided by an expert or generated by
the system, and the word ``class'' is used to refer to a single class in the
clustering.

\section{Evaluation Approach}

As mentioned above, we intend to be able to compare a clustering generated by
a system against one provided by an expert.  Since a word can occur in more
than one class, it is important to find some kind of mapping between the
classes generated by the system and the classes given by the expert.  Such a
mapping tells us which class in the system's clustering maps to which one in
the expert's clustering, and an overall comparison of the clusterings is based
on the comparison of the mutually mapping classes.

\begin{table}
   \caption{Two Example Classes} \vspace{8pt} \centerline{
   \begin{tabular}{|l|l|} \hline Class A & Class B \\
   (System) & (Expert) \\ \hline \hline cat &
   horse \\ dog & cow \\ stomach & cat \\ pig & pig \\ cow & lamb \\ hair &
   dog \\ cattle & sheep \\ goat & mare \\ & cattle \\ & swine \\ & goat \\
   \hline \hline \end{tabular} \label{table:classes} }
\end{table}

Before we delve deeper into the evaluation process, we must decide on some
measure of ``closeness'' between a pair of classes.  We have adopted the
F-measure \cite{HatMcK-93,Chi-92}.  In our computation of the F-measure,
we construct a contingency table based on the presence or absence of
individual elements in the two classes being compared, as opposed to basing it
on pairs of words.  For example, suppose that Class~A is generated by the
system and Class~B is provided by an expert (as shown in Table~1).  The
contingency table obtained for this pair of classes is shown in Table~2.

\begin{table}
   \caption{Contingency Table for Classes A and B}
   \vspace{8pt}
   \centerline{
   \begin{tabular}{|l||c|c|} \hline
   & \multicolumn{2}{|c|}{Expert} \\ \cline{2-3}
   & YES & NO \\ \hline \hline
   System - YES & 6 & 2  \\ \hline
   System - NO  & 5 & 0  \\ \hline \hline
  \end{tabular}
  \label{table:contable}
  }
\end{table}

The three main steps in the evaluation process are the acquisition of
``correct'' classes from domain experts, mapping the experts' clustering to
that generated by the system, and generating an overall measure that
represents the system's performance when compared against the expert.

\subsection{Knowledge Acquisition from Experts}

The objective of this step is to get human experts to undertake the same task
that the system performs, i.e., classifying a set of words into several
potentially overlapping classes.  The classes produced by a system are later
compared to these ``correct'' classifications provided by the expert.

\subsection{Mapping Algorithm}

In order to determine pairwise mappings between the clustering generated by
the system and one provided by an expert, a table of F-measures is
constructed, with a row for each class generated by the system, and a column
for every class provided by the expert.  Note that since the expert actually
provides a hierarchy, there is one column corresponding to every individual
class and subclass provided by the expert.  This allows the system's classes
to map to a class at any level in the expert's hierarchy.  This table gives an
estimate of how well each class generated by the system maps to the ones
provided by the expert.

The algorithm used to compute the actual mappings from the F-measure table is
briefly described here.  In each row of the table, mark the cell with the
highest F-measure as a potential mapping.  In general, conflicts arise when
more than one class generated by the system maps to a given class provided by
the expert.  In other words, whenever a column in the table has more than one
cell marked as a potential mapping, a conflict is said to exist.  To resolve a
conflict, one of the system classes must be re-mapped.  The heuristic used
here is that the class for which such a re-mapping results in minimal loss of
F-measure is the one that must be re-mapped.  Several such conflicts may
exist, and re-mapping may lead to further conflicts.  The mapping algorithm
iteratively searches for conflicts and resolves them till no more conflicts
exist.  Note also that a system class may map to an expert class only if the
F-measure between them exceeds a certain threshold value.  This ensures that a
certain degree of similarity must exist between two classes for them to map to
each other.  We have used a threshold value of 0.20.  This value is obtained
purely by observations made on the F-measures between different pairs of
classes with varying degrees of similarity.

\subsection{Computation of the Overall F-measure}

Once the mappings have been determined between the clusterings of the system
and the expert, the next step is to compute the F-measure between the two
clusterings.  Rather than populating separate contingency tables for every
pair of classes, construct a single contingency table.  For every pairwise
mapping found for the classes in these two clusterings, populate the YES-YES,
YES-NO, and NO-YES cells of the contingency table appropriately (see Table~2).
Once all the mapped classes have been incorporated into this contingency
table, add every element of all unmapped classes generated by the system to
the YES-NO cell and every element of all unmapped classes provided by the
expert to the NO-YES cell of this table.  Once all classes in the two
clusterings have been accounted for, calculate the precision, recall, and
F-measure as explained in \cite{HatMcK-93}.

\section{Results and Discussion}

In one of our experiments, the 400 most frequent nouns in the Merck Veterinary
Manual were clustered.  Three experts were used to evaluate the generated noun
clusters.  Some examples of the classes that were generated by the system for
the veterinary medicine domain are PROBLEM, TREATMENT, ORGAN, DIET, ANIMAL,
MEASUREMENT, PROCESS, and so on.  The results obtained by comparing these noun
classes to the clusterings provided by three different experts are shown in
Table~3.  We have also experimented with the use of WordNet to improve the
classes obtained by a distributional technique.  Some initial experiments have
shown that WordNet consistently improves the F-measures for these noun classes
by about 0.05 on an average.  Details of these experiments can be found in
\cite{Aga-95}.

\begin{table}
   \caption{Noun Clustering Results}
   \vspace{12pt}
   \centerline{
   \begin{tabular}{|c||c|c|c|} \hline
   & \multicolumn{3}{|c|}{System} \\ \cline{2-4}
   \raisebox{1.5ex}[0cm][0cm]{Expert}
   & Precision & Recall & F-measure \\ \hline \hline
   Expert A & 75.38 & 29.09 & 0.42  \\ \hline
   Expert B & 77.08 & 25.23 & 0.38  \\ \hline
   Expert C & 73.85 & 37.88 & 0.50  \\ \hline \hline
  \end{tabular}
  \label{table:table-title}
  }
\end{table}

It is our belief that the evaluation scheme presented in this paper is useful
for comparing different clusterings produced by the same system or those
produced by different systems against one provided by an expert.  The
resulting precision, recall, and F-measure should not be treated as a kind of
``gold standard'' to represent the quality of these classes in some absolute
sense.  It has been our experience that, as semantic clustering is a highly
subjective task, evaluating a given clustering against different experts may
yield numbers that vary considerably.  However, when different clusterings
generated by a system are compared against the same expert (or the same set of
experts), such {\underline{relative}} comparisons are useful.

The evaluation scheme presented here still suffers from one major limitation
--- it is not capable of evaluating a hierarchy generated by a system against
one provided by an expert.  Such evaluations get complicated because of the
restriction of one-to-one mapping.  More work definitely needs to be done in
this area.

\end{document}